\documentclass[conference]{IEEEtran}
\usepackage{cite}
\usepackage{amsmath,amssymb,amsfonts}
\usepackage{algorithmic}
\usepackage{graphicx}
\usepackage{textcomp}
\usepackage{xcolor}
\usepackage{float}
\def\BibTeX{{\rm B\kern-.05em{\sc i\kern-.025em b}\kern-.08em
    T\kern-.1667em\lower.7ex\hbox{E}\kern-.125emX}}
\begin{document}

\title{Exoplanet Detection Using Machine Learning Models Trained on Synthetic Light Curves\\
\thanks{}
}

\author{\IEEEauthorblockN{Ethan Lo}
\IEEEauthorblockA{\textit{}\\
\textit{Walton High School}\\
Marietta, GA, USA \\
elo9908@gmail.com}
\and
\IEEEauthorblockN{Dan Chia-Tien Lo}
\IEEEauthorblockA{\textit{Department of Computer Science} \\
\textit{Kennesaw State University}\\
Marietta, GA, USA \\
dlo2@kennesaw.edu}
}

\maketitle

\begin{abstract}
With manual searching processes, the rate at which scientists and astronomers discover exoplanets is slow because of inefficiencies that require an extensive time of laborious inspections. In fact, as of now there have been about only 5,000 confirmed exoplanets since the late 1900s. Recently, machine learning (ML) has proven to be extremely valuable and efficient in various fields, capable of processing massive amounts of data in addition to increasing its accuracy by learning. Though ML models for discovering exoplanets owned by large corporations (e.g. NASA) exist already, they largely depend on complex algorithms and supercomputers. In an effort to reduce such complexities, in this paper, we report the results and potential benefits of various, well-known ML models in the discovery and validation of extrasolar planets. The ML models that are examined in this study include logistic regression, k-nearest neighbors, and random forest. The dataset on which the models train and predict is acquired from NASA's Kepler space telescope. The initial results show promising scores for each model. However, potential biases and dataset imbalances necessitate the use of data augmentation techniques to further ensure fairer predictions and improved generalization. This study concludes that, in the context of searching for exoplanets, data augmentation techniques significantly improve the recall and precision, while the accuracy varies for each model.
\end{abstract}


\section{Introduction}
With over 100 billion stars in the Milky Way Galaxy, astronomers assert that each star has at least one extrasolar planet, also called exoplanets \cite{Seager_2014}. The search for exoplanets, planets orbiting a star outside the solar system, has gained momentum with the introduction of machine learning (ML) in 2020 \cite{armstrong_2020} and the use of data analysis techniques in this field. However, the National Aeronautics Space Administration (NASA) has confirmed only about 5,000 exoplanets since the 1990s \cite{NASA_nd}. Some of the most innovative missions that have contributed to the discovery of currently known exoplanets to date include Kepler \cite{Borucki_2016}, K2 \cite{Howell_2014}, and the Transiting Exoplanet Survey Satellite (TESS) \cite{Ricker_2014}, which is still currently operating and may potentially yield the greatest output yet. These telescopes rely on complex mathematical and ML algorithms to span the vast sea of stars. The Validation of Exoplanet Signals using a Probabilistic Algorithm (VESPA) is a common technique before its retirement to validate exoplanet transit signals \cite{armstrong_2020}. It has now been superseded by the TRICERATOPS tool, which has been developed to be significantly more advanced than its precursors. However, TRICERATOPS still exhibits limitations, such as inflexibility in transit timing variations and false positive misclassifications. Nonetheless, the ability to effectively discover and research exoplanets allows scientists to ascertain how planets like Earth have formed and expand current knowledge of the universe.

\subsection{Exoplanet Discovery Process}
\begin{figure}[H]
    \centering
    \includegraphics[width = \columnwidth]{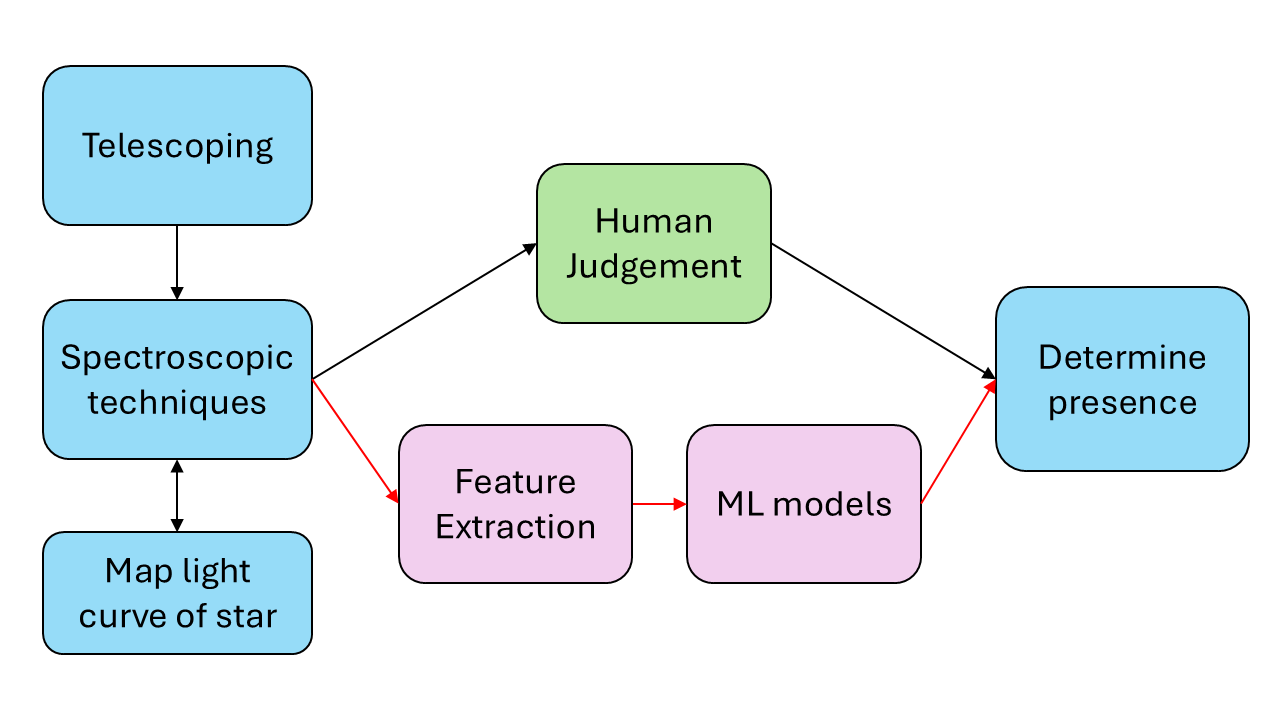}
    \caption{Diagram of Exoplanet Discovery Process}
    \label{fig:exoplanet_process_diagram}
\end{figure}

An exoplanet is confirmed when multiple observation methods verify its existence, whereas validation assesses its likelihood of being a genuine exoplanet \cite{Hoover_2021}. This paper focuses on the former confirmation. The process for the discovery of an exoplanet is as follows: first, researchers span the sky for particular stars through large, advanced telescopes. Then, through transmission spectroscopic techniques, the light curve of the star is developed by its relative brightness \cite{Seager_2014}. Astronomers use indirect spectroscopic methods to find exoplanets, since exoplanets barely reflect light from its star. A productive method is transit photometry, which measures the brightness of a star and how it dips when a planet crosses in front of it \cite{Charbonneau_2000}. Humans then judge the generated light curve to ascertain the validation of an exoplanet and later its confirmation. Alternatively, ML models can directly evaluate light curve data to detect planetary presence. A diagram of the process is displayed in Figure \ref{fig:exoplanet_process_diagram}.

\subsection{Reasons for Inefficiency}
Relying solely on automated ML models is not yet considered ideal, and it is thus suggested to include various other methods for confirmation. For example, the CoRoT satellite classified only light curves with an 83\% false positive rate \cite{Almenara_2009}. However, including numerous confirmation methods introduces many more uncertainties and complications. Due to current limited technology, the classification of such planets may be furthermore inaccurate from several factors. Binary systems or asteroids on the same trajectory, for instance, can warp observations \cite{Sekhar_2023}. Poisson noise may appear in the light curves, dependent on the accuracy of the sensor, blurring any apparent transit-like signals \cite{alvarez_2023}. Large, advanced telescopes, which are widely used for detection, can be extremely costly for maintenance and operation. For example, the TESS mission launched by NASA had a \$230 million cost cap exclusive of launch costs \cite{Seager_2014}. Even though advanced telescopes yield promising results in a much more efficient manner, the costly construction and intensive operation usage diminish its viability as a sustainable solution. Ultimately, with a low range of confirmed exoplanets, it remains difficult to effectively train these automated processes on a diverse set of real-world cases.

\subsection{Research Question}
Machine learning, a subset of artificial intelligence focused on learning tasks from pattern recognition \cite{Sun_2024}, has recently been introduced into this field. The use of ML algorithms easily overcomes the manual, labor-intensive, interpretation of light curves. ML algorithms enable astronomers to efficiently sift through vast datasets to identify potential exoplanet candidates. Additionally, ML algorithms thrive through many forms, with some outperforming others in certain situations. There exists a multitude of ML models available, but this particular situation of detecting exoplanets from flux data limits the context-suitability for the model. With ML algorithms, data augmentation techniques play a significant role in improving feature processing for model training. Normalization and filtering name just a few of the many data augmentation techniques possible. To differentiate genuine exoplanets from false cases, this paper explores the application of synthetic data generation on the accuracy, recall, and precision of three ML models in exoplanet detection: logistic regression, k-nearest neighbors, and random forest. Data augmentation techniques are planned to assist in balancing the data to improve model performance. To improve efficiencies while minimizing costs of operation, this paper seeks to determine what common, well-known ML models can most effectively detect the presence of true exoplanets with synthetic sampling techniques.

\section{Related Work}
Since the introduction of machine learning in the detection of exoplanets, models have been continuously updated to be more accurate and precise. On a supercomputer at NASA, researchers have developed ExoMiner, considered to be the most accurate ML model for the validation of exoplanets \cite{Hoover_2021}. ExoMiner relies on a deep learning neural network composed of layers of neurons. It has the ability to learn from false positive cases to correctly identify rare cases that consist of irregularities. The model can additionally provide reasoning for its conclusion of false and true positives and negatives. ExoMiner however relies on substantial, high-quality, labeled data and requires significant computational power, including specialized hardware like GPUs or TPUs.

Another neural network tested by professors in Spain in 2024 is a 1-dimensional convolutional neural network (CNN) \cite{alvarez_2023}. CNNs mimic the human brain with layers composed of artificial neurons through a convolution function. The 1D CNN model is trained to learn the shape of such light curves, consisting of transits, transits with noise, and pure noise. The computational time compared to other models is 0.1\% faster than box least squares (BLS), a traditional transit search method, with a negligible change in accuracy. The tested CNN model performed with an accuracy of 99.02\% and an estimated error of 0.03 \cite{alvarez_2023}. With a comparatively short training time, the model tested 300,000 different light curves, which would have taken an extensive amount of time if done by humans.

The detection of exoplanets can be complemented with machine learning in object detection. According to the International Business Machines Corporation (IBM), object detection is a computer vision task that finds and classifies objects in images through neural networks \cite{Murel_2024}. Its applications are limitless, ranging from security imaging to autonomous driving. Object detection is a more specific machine learning model in that it utilizes a convolutional neural network structure. A research study using the third and fifth versions of the object detection model You Only Look Once (YOLO) determined its precision in detecting dips in an exoplanet’s transit through a light curve \cite{Sekhar_2023}. Their YOLO V5 model had a precision of 0.856 – 15\% better than other well-known object detection models \cite{Sekhar_2023}. The resulting precision is not as high as expected in relation to the CNN discussed earlier. However, YOLO has some advantageous features over CNNs in that the training YOLO models takes less time and can be run on resource-limited devices, decreasing the computational intensity.

\section{Dataset}
\begin{figure}[H]
    \centering
    \includegraphics[width = \columnwidth]{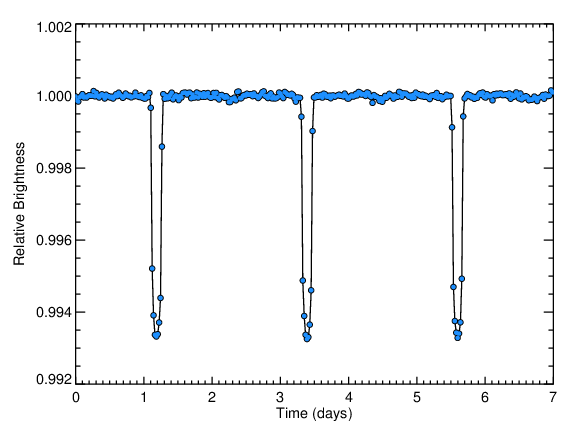}
    \caption{Example of a Light Curve of an Exoplanet}
    \label{fig:light_curve_example}
\end{figure}

This project used data from NASA’s Kepler space telescope with over 30 years of information \cite{WDelta_2017}. With a shape of 5,087 by 3,197, the dataset is comprised of 5,050 non-exoplanets and 37 exoplanets, where exoplanets comprise an extremely low 0.73\% of all cases. The columns represent flux data, the light intensity of the star's relative brightness. Each star has a light curve, the relative brightness over a period of time. For simplicity's sake, light curves will be differentiated by referring to light curves of an exoplanet and a non-exoplanet. When a planet orbits in front of its star, the star's relative brightness ever so slightly decreases. The transit of an exoplanet is identified by a dip in the relative brightness as it transits past Earth's perspective. A graph of a light curve displaying the presence of an exoplanet is shown in Figure \ref{fig:light_curve_example}. 

\begin{figure}[H]
    \centering
    \includegraphics[width = \columnwidth]{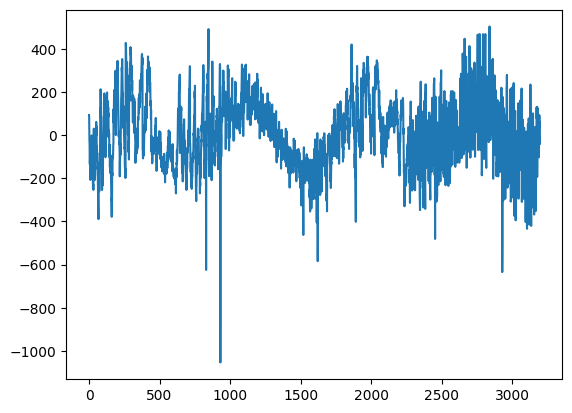}
    \caption{Visualization of a Light Curve of an Exoplanet from Dataset}
    \label{fig:light_curve_dataset}
\end{figure}

Regarding the universe, exoplanet transits are not as easily identifiable as in Figure \ref{fig:light_curve_example}. In the dataset used for this project, the light curves of exoplanets and non-exoplanets can be nearly indistinguishable. Figure \ref{fig:light_curve_dataset} shows a more practical depiction of a light curve of an exoplanet.

\begin{figure}[H]
    \centering
    \includegraphics[width = \columnwidth]{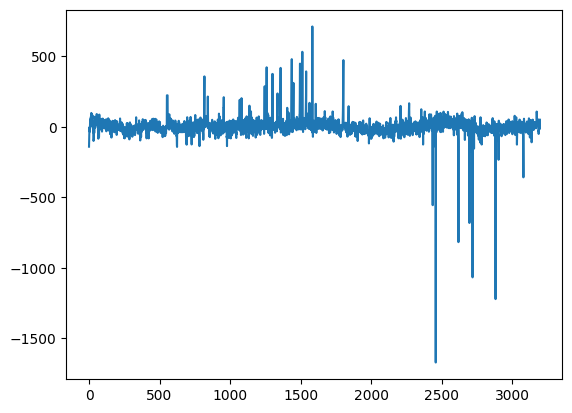}
    \caption{Visualization of a Light Curve of a Non-exoplanet from Dataset}
    \label{fig:light_curve_nonexoplanet_dataset}
\end{figure}

As shown in Figure \ref{fig:light_curve_dataset}, there is somewhat of a period that is maintained throughout the observed time. In contrast, a non-exoplanet light curve as in Figure \ref{fig:light_curve_nonexoplanet_dataset} differs because there is no consistent pattern. 

The machine learning models essentially examine the flux data and identify any present transits to determine whether a candidate is an exoplanet.

\section{Methods}
This section introduces and examines the proposed machine learning models: logistic regression, k-nearest neighbors (KNN), and random forest, imported from the scikit-learn library \cite{scikit_learn}. The dataset is divided into 80\% for training and 20\% for testing. In the training data, 26 are exoplanets while 4043 are non-exoplanets, and in the testing data, 11 are exoplanets while 1007 are non-exoplanets.

\subsection{Logistic Regression}
Logistic regression is a statistical model used for binary classification problems, as portrayed by labeling an exoplanet as one and a non-exoplanet as zero. It uses the Sigmoid function in equation \ref{eq:sigmoid}, a logistic function that takes in feature values to output a probability of an outcome. 

\begin{equation}
f(x)=\frac{1}{1+e^{-x}}
\label{eq:sigmoid}
\end{equation}

\begin{figure}[H]
    \centering
    \includegraphics[width = \columnwidth]{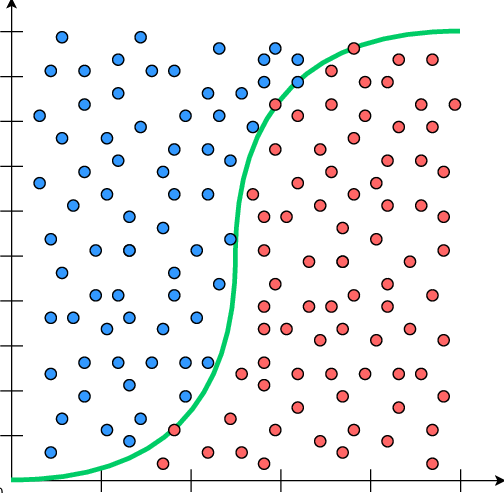}
    \caption{Example of Logistic Regression}
    \label{fig:logistic-regression-example}
\end{figure}

In exoplanet detection, it predicts whether a given set of features corresponds to an exoplanet or not. The model assumes a linear relationship between the input features and the log-odds of the outcome. Despite its simplicity, logistic regression is effective when the relationship between features and the target variable is approximately binary. However, it may struggle with complex, non-linear relationships that are common in astronomical data. In non-linear relationships with two classes, logistic regression can clearly distinguish them as seen in Figure \ref{fig:logistic-regression-example} \cite{logistic-regression-example}.

\subsection{K-Nearest Neighbors (KNN)}
\begin{figure}[H]
    \centering
    \includegraphics[width = \columnwidth]{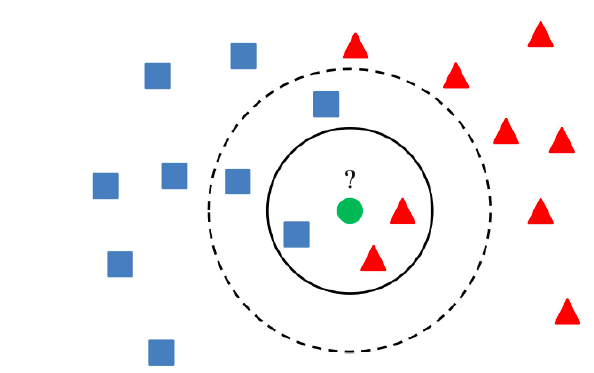}
    \caption{Example of KNN}
    \label{fig:knn-diagram}
\end{figure}

K-Nearest Neighbors (KNN) is a ML algorithm that classifies a data point based on the majority class of its k-nearest neighbors in the feature space, exemplified in Figure \ref{fig:knn-diagram} \cite{knn-diagram}. The k value is variable and determines how many neighbors should the model consider to classify the data point. In the context of exoplanet detection, KNN can be used to classify a star's light curve data as either containing an exoplanet signal or not. The choice of k is crucial; a small k may lead to false labeling, while a large k could smooth out critical patterns yet overestimate situations. Figure \ref{fig:knn-diagram} shows an example usage of KNN with three and five neighbors.

\subsection{Random Forest}
\begin{figure}[H]
    \centering
    \includegraphics[width = \columnwidth]{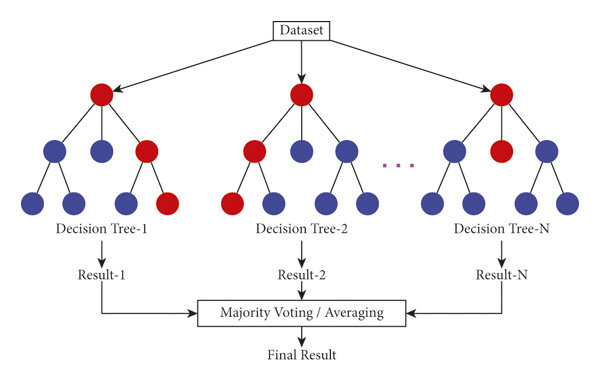}
    \caption{Diagram of Random Forest Process}
    \label{fig:rf-diagram}
\end{figure}

The random forest algorithm is an ensemble method, which is made up of a set of classifiers as shown in Figure \ref{fig:rf-diagram} \cite{random_forest}. In this case, a random forest consists of multiple decision trees and combines the output of each to reach a single conclusion. Decision trees are versatile and interpretable models that divide the data based on feature values to make predictions. For exoplanet detection, random forests can effectively capture non-linear relationships and interactions between features. While a decision tree can be prone to overfitting, random forest effectively eliminates that issue. 

\section{Preliminary Results of Proposed ML Models}
This section presents and discusses the implications behind the preliminary results of the three proposed ML models. After specialized analyses of appropriate parameters, each model is trained and tested to analyze their accuracy in classifying exoplanets. Their accuracy is evaluated numerically and visually with a confusion matrix.

\subsection{Logistic Regression Model Results}
\begin{figure}[H]
    \centering
    \includegraphics[width = \columnwidth]{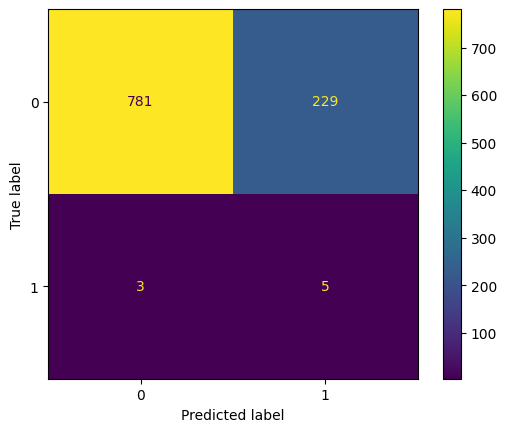}
    \caption{Confusion Matrix of Tested Logistic Regression Model}
    \label{fig:log_reg_cfm}
\end{figure}

Using the logistic regression classifier from the scikit-learn library \cite{scikit_learn}, the maximum number of iterations for solvers to converge is set to 1000. Figure \ref{fig:log_reg_cfm} depicts the results of the model in a confusion matrix. A value of one represents the presence of an exoplanet and a value of zero a non-exoplanet. The model trains with an accuracy of 93.9\% on 80\% of the data. However, when comparing the model’s predictions in the remaining 20\% of the data with their corresponding values, its accuracy drops to 77.2\% on the test data. The recall rate is calculated to be 62.5\% and the precision is 2.1\%, which are quite low.

As seen in Figure \ref{fig:log_reg_cfm}, the logistic regression model has a large bias towards exoplanets. The model produces 229 false positives with a low accuracy score when tested, which may be partially due to overfitting in binary classification and thus poor prediction performance on the minority class.

\subsection{K-Nearest Neighbors Model Results}
\begin{figure}[H]
    \centering
    \includegraphics[width = \columnwidth]{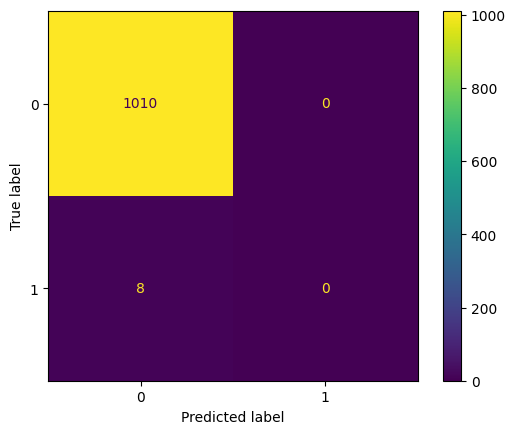}
    \caption{Confusion Matrix of Tested KNN Model}
    \label{fig:knn_cfm}
\end{figure}

Imported from the scikit-learn library \cite{scikit_learn}, the KNN classifier is trained and tested on the same dataset. With a k value of 4 nearest neighbors, the KNN model trains with an accuracy of 99.3\% and predicts with an accuracy of 99.2\%. This accuracy, though it seems excellent by itself, may have incorporated potential biases. As seen in Figure \ref{fig:knn_cfm}, the model favors non-exoplanets more because exoplanets consist of fewer than 0.1\% of the entire data. The calculated recall rate and precision for KNN are both 0\%. Though all non-exoplanets are correctly classified, disappointingly, the KNN model predicts no correct exoplanets.

\subsection{Random Forest Model Results}
\begin{figure}[H]
    \centering
    \includegraphics[width = \columnwidth]{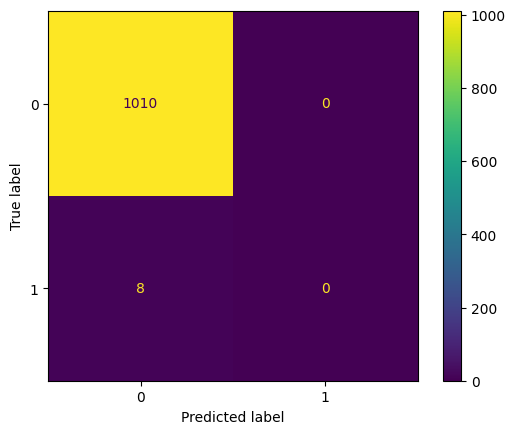}
    \caption{Confusion Matrix of Tested Random Forest Model}
    \label{fig:random_forest_cfm}
\end{figure}

The selected random forest model is also imported from the scikit-learn library \cite{scikit_learn}. The parameters selected are 250 trees in the forest at a random state of 0. With the same dataset, the random forest classifier model trains with a 100\% accuracy and performs with a 99.2\% on the test data as shown in Figure \ref{fig:random_forest_cfm}, yet it likewise predicts zero true positives. The recall rate and precision are again 0\%.

\subsection{Implications}
Indeed, there are much too many false positive predictions by each ML model, suggesting a problem with the dataset. When there is not enough data for each classification (one or zero), the model provides its own bias based on its training. The minority class may lack enough examples for the model to learn meaningful patterns. When the dataset contains significantly more examples of one class (the majority class) than others, the model learns to predict the majority class more frequently. During training, the model might overfit to the dominant class, ignoring rare patterns present in the minority class. This minimizes the overall error, and the undesirable output may not be clearly evident unless some visual depiction, like a confusion matrix, is displayed.

\section{Data Augmentation Techniques}
One challenge in exoplanet detection is the imbalance in datasets, where the number of positive exoplanet cases is significantly smaller than the number of negative cases. Because of the lack of confirmed exoplanets for training, simulated light curves and human-vetted Kepler TCEs can provide additional training for a more reliable accuracy \cite{alvarez_2023}. This project uses data augmentation techniques to generate more flux data. In this case, all exoplanets in the dataset represent less than 0.1\%, which leads to biased models that favor the majority class. To address this issue, five data augmentation techniques are utilized: Fourier-based augmentation, Savitzky-Golay filter, normalization, RobustScalar augmentation, and Synthetic Minority Oversampling Technique (SMOTE). Fourier-based augmentation enhances the robustness of ML models against common corruptions by training them on augmented datasets \cite{Mehr_2022}. The Savitzky-Golay filter is a technique used to reduce noise while preserving the shape and features of the original signal, such as peaks and edges. To calculate convolution weights at all positions of the data, the filter fits a low-degree polynomial to a moving window of the data via least squares \cite{Gorry_1990}. Normalization scales the data so that all flux data lie between zero and one. The RobustScalar method is a preprocessing technique for scaling features by removing the median and scaling them based on the interquartile range (IQR), making the data robust to outliers \cite{RobustScalar}. 

\subsection{Synethic Sample Generation}
\begin{figure}[H]
    \centering
    \includegraphics[width = \columnwidth]{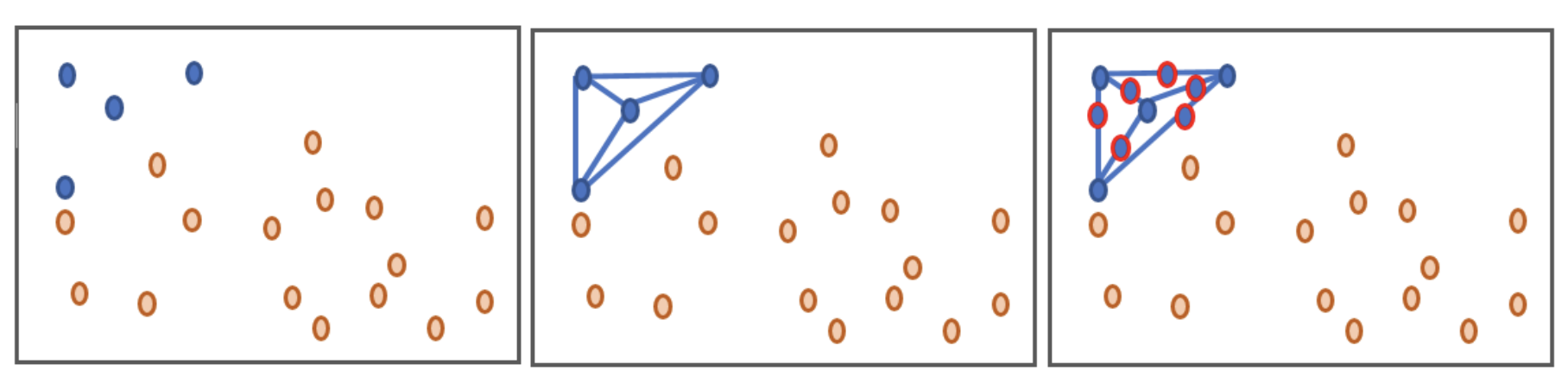}
    \caption{Visualization of SMOTE Augmentation}
    \label{fig:smote}
\end{figure}

SMOTE generates synthetic samples for the minority class by interpolating between existing minority samples and their nearest neighbors to better define classification regions \cite{Kummer_2022}. Because there are very few instances of true exoplanets in the dataset, SMOTE would be the most pivotal technique in balancing the dataset by adding true values of exoplanets to the dataset and helping the model learn to detect true exoplanets more effectively. Figure \ref{fig:smote} visually simplifies the process of SMOTE augmentation.

\begin{figure}[H]
    \centering
    \includegraphics[width = \columnwidth]{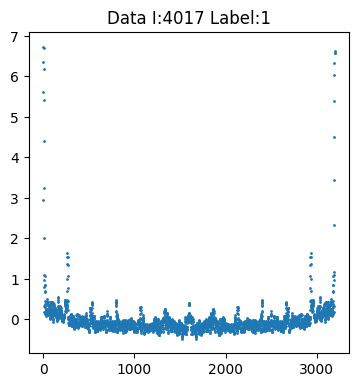}
    \includegraphics[width = \columnwidth]{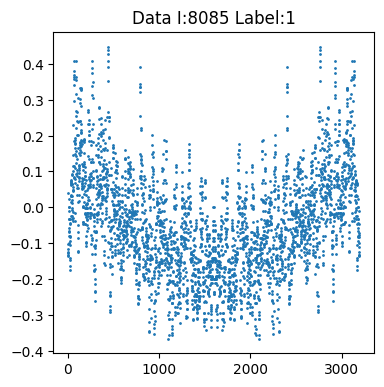}
    \caption{Visualization of Original vs. SMOTE Augmented Exoplanet Data}
    \label{fig:smote_comparison}
\end{figure}

SMOTE only affects the data with exoplanets, as they are the minority class; there is no need to generalize the majority class data. The exoplanet data augmented with SMOTE consist of many more data values, which helps to balance the data as a whole. As seen in Figure \ref{fig:smote_comparison}, SMOTE generates many more samples containing exoplanets, indicated by their index value. 

\section{Results and Discussion After Data Augmentation}
After implementing the data augmentation techniques, the dataset now consists of 10,100 data points, where 5,050 points represent non-exoplanets and the other 5,050 exoplanets, which should minimize any issues with bias and overfitting. A possible limitation, however, may be that because there are only 37 exoplanet cases to begin with, the generation of other exoplanets are dependent on a limited pool of variability. Training and testing is again split into an 80-20 ratio. In the training data, there are 4043 exoplanets and non-exoplanets each, while in the testing data, there are 1007 exoplanets and non-exoplanets each. All parameters for each ML model remain the same in order to compare the effects of data augmentation.

\subsection{Augmented Logistic Regression Model Results}
Having trained with a 100\% accuracy, the augmented logistic regression model performs with an accuracy of 91.0\%. The impact of the augmented data resulted in a 13.8 percent increase in model performance. The calculated recall is now 83.1\% and the precision is 98.7\%.

\begin{figure}[H]
    \centering
    \includegraphics[width = \columnwidth]{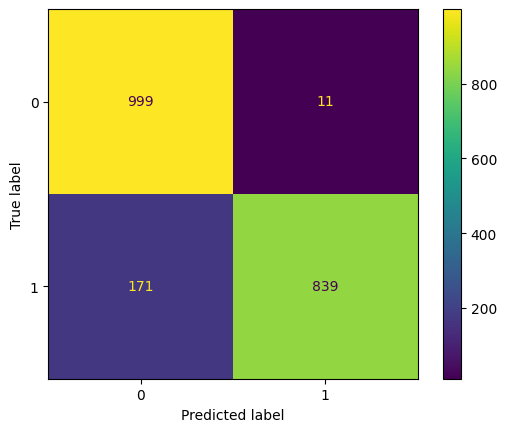}
    \caption{Confusion Matrix of Tested Augmented Logistic Regression}
    \label{fig:aug_lr_cfm}
\end{figure}

After data augmentation, the confusion matrix highlights issues with false negatives, contrasting the earlier false positive bias. Nonetheless, the logistic regression model predicts comparatively fairer between exoplanets and non-exoplanets. A 20.6 percent increase and 96.6 percent increase is seen in the recall and precision, both considerable changes of improvement.

\subsection{Augmented KNN Model Results}
\begin{figure}[H]
    \centering
    \includegraphics[width = \columnwidth]{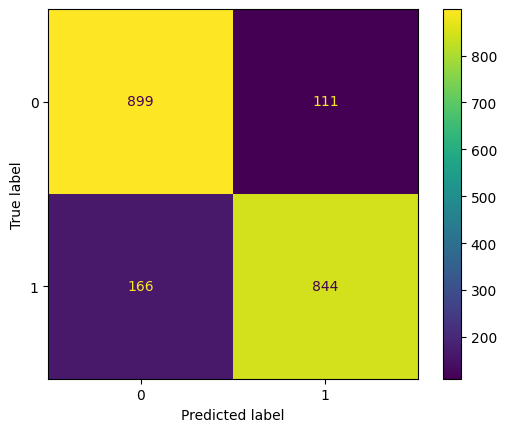}
    \caption{Confusion Matrix of Tested Augmented KNN}
    \label{fig:aug_knn_cfm}
\end{figure}

Just as before, the k value is set to 4 nearest neighbors. After data augmentation techniques, the KNN model's accuracy decreases in both training and testing to 96.7\% and 86.3\%, respectively. The number of false positives and false negatives is relatively similar as shown in Figure \ref{fig:aug_knn_cfm}, which implies that the model did not have a considerable bias towards either class. The calculated recall for the augmented KNN is 83.6\% and the precision is now 88.4\%.

\subsection{Augmented Random Forest Model Results}
With the same 250 trees at a random state of 0, the random forest model after data augmentation has a training accuracy of 100\% and a testing accuracy of 87.3\%. A decrease of 11.9\% is seen, yet a relatively decent accuracy is still retained. 

\begin{figure}[H]
    \centering
    \includegraphics[width = \columnwidth]{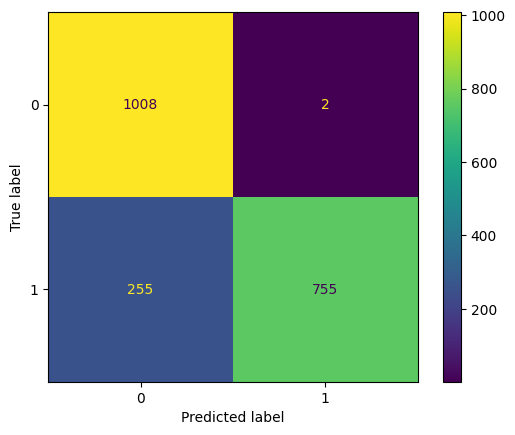}
    \caption{Confusion Matrix of Tested Augmented KNN}
    \label{fig:aug_rf_cfm}
\end{figure}

The random forest model only predicts two false positives but 255 false negatives. The calculated recall of the augmented random forest is 74.8\% and the precision is now 99.7\%. The model seems to be more inclined to the non-exoplanet class rather than the exoplanet class, which indicates a bias towards the non-exoplanet class still.

\subsection{Comparison of Related Work, Preliminary, and Augmented Results}
Although decreases in accuracy are evident for each ML model except logistic regression after data augmentation techniques are employed, the recall and precision values greatly increase. The percentage change in accuracy for each model is mentioned previously. For each model after data augmentation, logistic regression (LR) has a 20.6\% increase and 96.6\% increase; KNN has a 83.6\% increase and 88.4\% increase; and random forest (RF) has a 74.8\% increase and 99.7\% increase in recall and precision, respectively. In comparison to the ExoMiner model, which has a precision of 99\% \cite{Valizadegan_2022}, the augmented RF yields a higher precision of 99.7\%. However, the augmented RF has a 74.8\% recall. Although the recall and precision of YOLO are considerably lower than most of those of other models, the tested YOLO model is run on low computing power devices, such as Raspberry Pi, mobile phones, and other microcomputers \cite{Sekhar_2023}. The CNN tested by researchers in Spain has an accuracy of 99.0\%, though the recall and precision are not reported \cite{alvarez_2023}. 

\subsubsection{Evaluating the F1-score}

The F1-score takes the harmonic mean of both the recall and precision. Recall is sensitive to false positives as precision is to false negatives, and when exoplanets need to be distinguished from both false negatives and positives, the F1-score nicely considers both weightings. The F1-scores before augmentation for LR, KNN, and RF are 4.1\%, 0\%, and 0\% respectively. After augmentation, the F1-scores for LR, KNN, and RF are all significantly increased to 90.2\%, 85.9\%, and 85.5\% respectively, where LR trumps the other models. A table of each model's results with the related work results is shown in Table \ref{tab:comparison-results}.

\begin{table}[H]
    \centering
    \begin{tabular}{|l|l|l|l|l|}
        \hline
        Model & Accuracy (Test) & Recall & Precision & F1-score\\ \hline
        LR & 77.2\% & 62.5\% & 2.1\% & 4.1\%\\ \hline
        KNN & 99.2\% & 0.0\% & 0.0\% & 0.0\%\\ \hline 
        RF & 99.2\% & 0.0\% & 0.0\% & 0.0\%\\ \hline
        Augmented LR & 91.0\% & 83.1\% & 98.7\% & 90.2\%\\ \hline
        Augmented KNN & 86.3\% & 83.6\% & 88.4\% & 85.9\%\\ \hline
        Augmented RF & 87.3\% & 74.8\% & 99.7\% & 85.5\%\\ \hline
        ExoMiner & 73.6\% & 93.6\% & 99\% & 96.2\%\\ \hline
        YOLO & - & 85\% & 81\% & 83.0\%\\ \hline
        CNN & 99.0\% & - & - & -\\ \hline
    \end{tabular}
    \caption{Comparison Among Related Work to Tested ML Models}
    \label{tab:comparison-results}
\end{table}

\section{Conclusion and Future Work}
The search for exoplanets using machine learning models and data augmentation techniques presents an innovative intersection of astronomy and computer science. Logistic regression, KNN, and random forest each offer unique strengths and weaknesses in this endeavor. Logistic regression provides simplicity and interpretability, KNN offers a non-parametric approach, and random forest captures complex feature interactions. With a dataset imbalance, issues of bias and overfitting can occur especially in exoplanet discovery, where there are already very few recorded instances of exoplanets. The effectiveness of these models can be further enhanced by addressing challenges such as data imbalance, overfitting, and feature scaling through data augmentation techniques. The augmentation techniques explored in this paper include Fourier-based augmentation \cite{Mehr_2022}, Savitzky-Golay filter \cite{Gorry_1990}, normalization, RobustScalar method \cite{RobustScalar}, and SMOTE \cite{Kummer_2022}. SMOTE is considerably the most pivotal technique in this study, enhancing the classification by providing more specific data instances that the model can analyze. Synthetic sampling in the context of exoplanet detection is crucial for better model performance during training. With regard to precision, the augmented RF yields the best result of 99.7\%. If recall is considered, the most suitable model appears to be the augmented KNN with a score of 83.6\%. The highest accuracy reported is 91.0\% after the augmentation on LR. However, the F1-score should be the most examined metric because it takes in the harmonic average of the recall and precision, two major insights in exoplanet detection. Thus, the model that performs best after data augmentation techniques are employed is logistic regression with a F1-score of 90.2\%.

To further improve the accuracy of the tested ML models, hyper-parameter tuning can be employed, though there are not many parameters needed for each model as each is a relatively simple, well-known algorithm. Other data augmentation techniques can be utilized to assist the ML models in training and predicting as well. Additionally, removing outliers is considered during the research of this project but is disregarded because no matter the standard deviation, all exoplanet data are removed because of its exceedingly low number before data augmentation. This paper offers the possibility of common, well-known ML models to comparatively perform with complex models, such as ExoMiner and other deep learning networks.

The understanding of otherworldly concepts in exoplanets can transform the field of space exploration in numerous areas, such as planet habitability, planet formation or composition, and much more. With planets millions of light-years away from Earth, characteristics of exoplanets, such as atmospheric conditions or composition, can be examined with advanced technology that will only continue to advance in form. Machine learning models process data not only efficiently but also at an unprecedented massive scale. Utilizing common, well-known ML models can minimize complexities while retaining efficiency. Other fields, such as healthcare or finance, can benefit from the use of simpler ML algorithms that reduce costs of operation and construction or additional complications. The implementation of these propositions allow for new discoveries and insights into the astronomical field. As technology advances and datasets grow, machine learning will continue to play a crucial role in the discovery of new worlds beyond the solar system.

\bibliographystyle{IEEEtran}
\bibliography{sample-base}
\end{document}